\documentclass[final]{cvpr}

\usepackage{times}
\usepackage{epsfig}
\usepackage{graphicx}
\usepackage{amsmath}
\usepackage{amssymb}

\usepackage{multirow}

\usepackage[pagebackref=true,breaklinks=true,colorlinks,bookmarks=false]{hyperref}

\def\cvprPaperID{****} 
\def\confYear{CVPR 2021}

\begin{document}

\title{Team VI-I2R Technical Report on EPIC-KITCHENS-100 Unsupervised Domain Adaptation Challenge for Action Recognition 2021}

\author{
Yi Cheng, Fen Fang, Ying Sun\\
Institute for Infocomm Research, A*STAR, Singapore\\
{\tt\small \{cheng\_yi, fang\_fen, suny\}@i2r.a-star.edu.sg}
}

\maketitle

\begin{abstract}
In this report, we present the technical details of our approach to the EPIC-KITCHENS-100 Unsupervised Domain Adaptation (UDA) Challenge for Action Recognition. The EPIC-KITCHENS-100 dataset consists of daily kitchen activities focusing on the interaction between human hands and their surrounding objects. It is very challenging to accurately recognize these fine-grained activities, due to the presence of distracting objects and visually similar action classes, especially in the unlabelled target domain. Based on an existing method for video domain adaptation, \ie, TA3N, we propose to learn hand-centric features by leveraging the hand bounding box information for UDA on fine-grained action recognition. This helps reduce the distraction from background as well as facilitate the learning of domain-invariant features. To achieve high quality hand localization, we adopt an uncertainty-aware domain adaptation network, \ie, MEAA, to train a domain-adaptive hand detector, which only uses very limited hand bounding box annotations in the source domain but can generalize well to the unlabelled target domain. Our submission achieved the 1st place in terms of top-1 action recognition accuracy, using only RGB and optical flow modalities as input. 

\end{abstract}

\section{Introduction}

\begin{figure}[ht]
     \centering
     \includegraphics[width=1.0\linewidth]{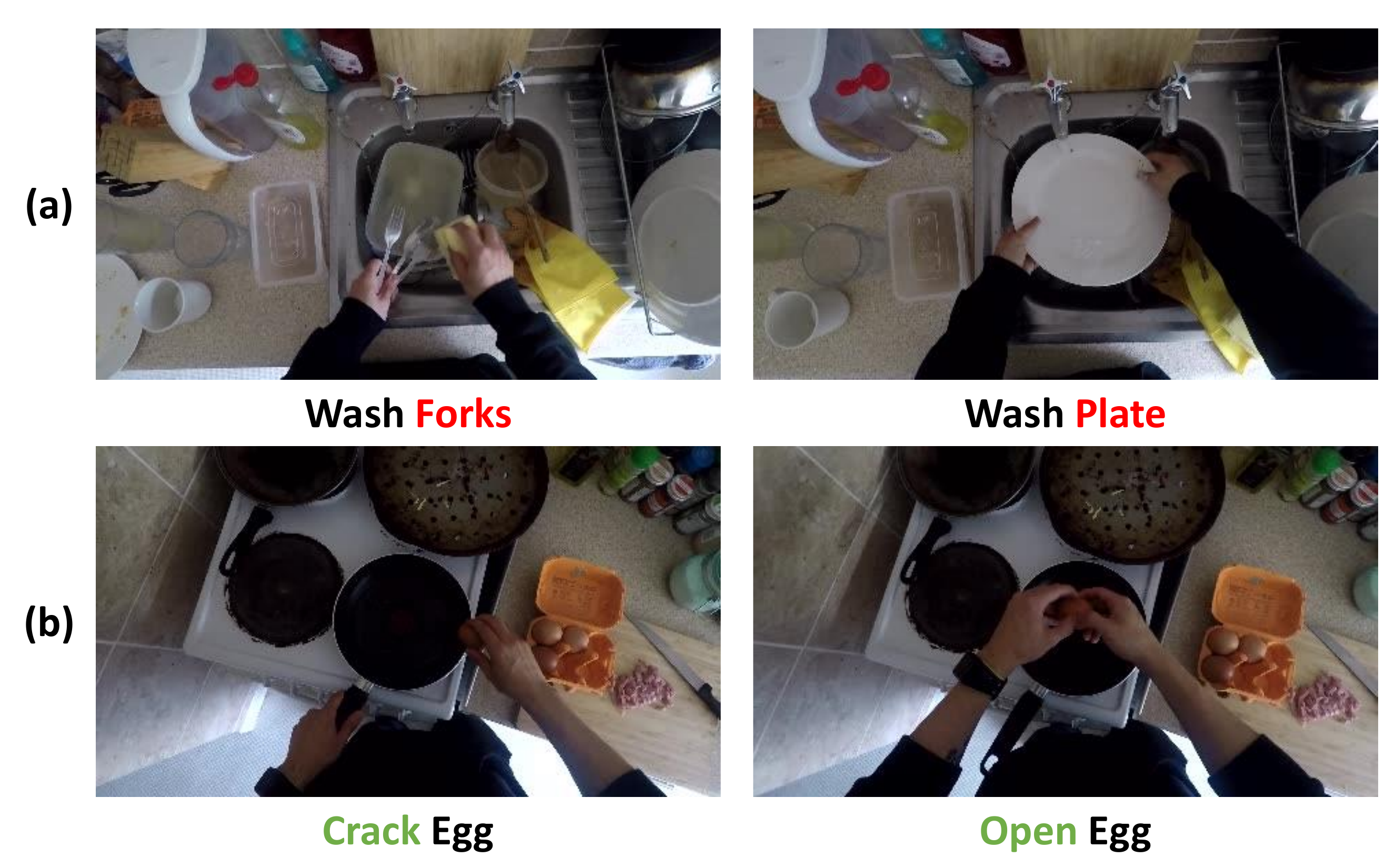}
     \caption{Illustration of the challenges in fine-grained action recognition on EPIC-KITCHENS-100 dataset. (a) Presence of distracting objects: there are many distracting objects in the scene, \eg, sponge and tap, which makes it difficult to identify the active objects. (b) Visually similar actions, \eg, ``crack egg'' versus ``open egg'', which requires to capture the subtle difference between hand motions. 
     To handle both of these two challenges, it is important to enhance the features around hand regions.
     }
     \label{figure:intro}
\end{figure}

The EPIC-KITCHENS-100 dataset is a large-scale video benchmark, capturing daily cooking activities from egocentric perspective~\cite{damen2020rescaling}. It mainly contains fine-grained actions which reflect the interaction between human hands and their surrounding objects, and each action class is defined by a verb and a noun class. The EPIC-KITCHENS-100 Unsupervised Domain Adaptation (UDA) Challenge for Action Recognition aims to adapt an action recognition model trained on a labelled source domain to an unlabelled target domain. In this challenge, the source domain contains egocentric videos captured in 2018, while the target domain contains egocentric videos captured in 2020 with the same subjects but different cameras and potential change of kitchens. This is a challenging task, as the source and target domains have different data distributions, due to the changes of environments and camera settings. Solutions successfully addressing this challenge can help save much time and efforts when applying the model trained on an existing labelled dataset to a newly collected dataset without annotation.

\begin{figure*}[t]
  \centering
  \includegraphics[width=1.0\linewidth]{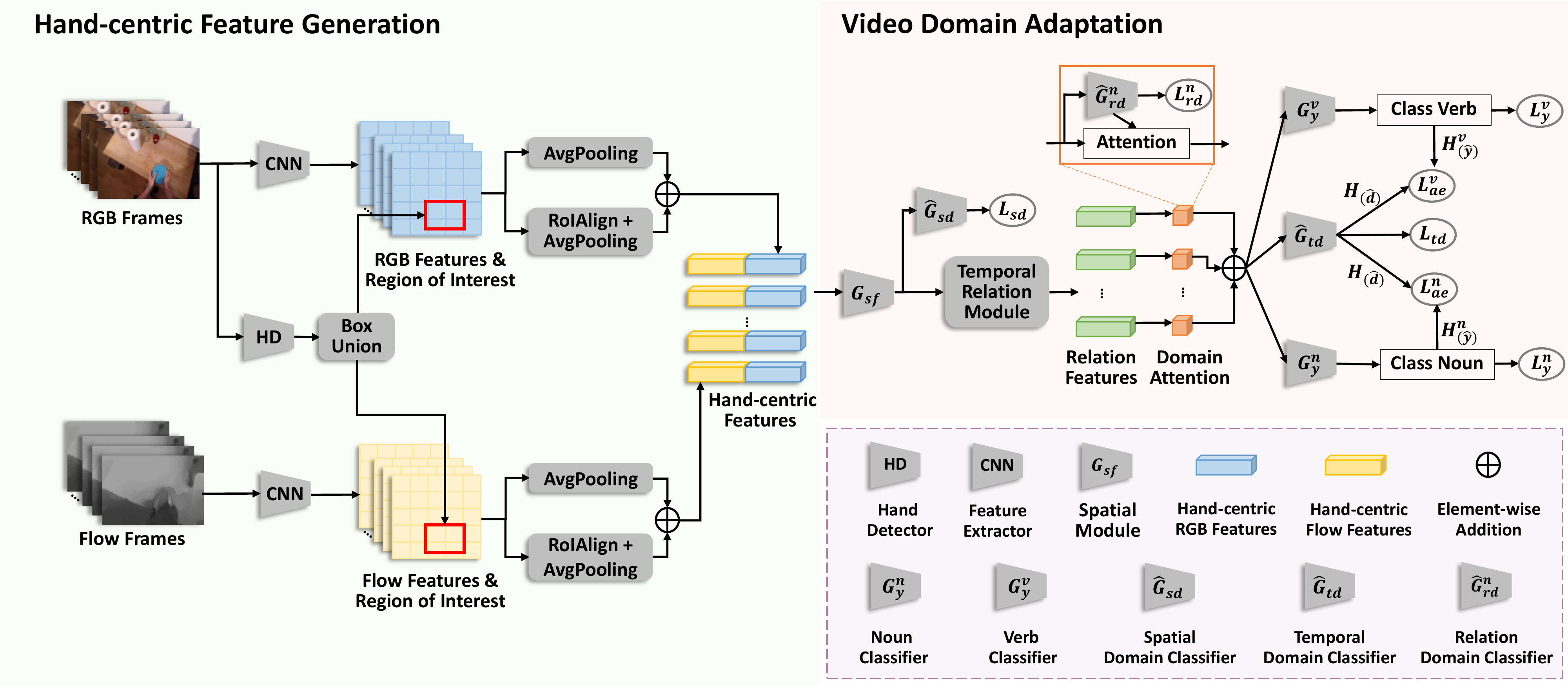}
  \caption{Overall architecture of the proposed framework. $L_{sd}$, $L_{td}$ and $L_{rd}^n$ denote the spatial, temporal and relation domain classification loss, respectively. $L_y^n$ and $L_y^v$ denote the noun and verb classification loss, respectively. $H(\hat{d})$, $H(\hat{y}^n)$ and $H(\hat{y}^v)$ denote the entropy of predictions from temporal domain classifier, noun classifier and verb classifier, respectively. $L_{ae}^n$ and $L_{ae}^v$ denote the attentive entropy loss for noun and verb, respectively.
  This is best viewed in color.} 
\label{fig:architect}
\end{figure*}
\vspace{3mm}

In Fig~\ref{figure:intro}, we present some samples from the EPIC-KITCHENS-100 dataset. As shown in the figure, activities in this dataset mainly focus on the interactions between human hands and their surrounding objects. This brings two challenges for fine-grained action recognition. The first challenge is the presence of distracting objects in the scene, which makes it difficult to identify the active objects. Based on the observation that active objects are generally located around the hand regions, we find that it is important to enhance the features around the hand regions~\cite{fan2021understanding}. The other challenge is that some action classes are very similar visually, where capturing the subtle difference among hand motions are essential to accurately recognize the target actions. Therefore, enhancing the features around hand regions can help to reduce the distraction from background and thus improving the action recognition accuracy. Moreover, this may provide further benefits in the context of UDA for fine-grained action recognition by facilitating the learning of domain-invariant features. To the best of our knowledge, MM-SADA~\cite{munro2020multi} is the first attempt on UDA for fine-grained action recognition. It leverages the multi-modal nature of video data to adapt fine-grained action recognition models to unlabelled target domain, which provides the first benchmark on this task. However, it does not consider the aforementioned challenges.

To address these two challenges, we propose to learn hand-centric features by leveraging the hand bounding box information for UDA on fine-grained action recognition. Specifically, we adapt the TA3N~\cite{chen2019temporal}, an existing method for video domain adaptation, to learn a fine-grained action recognition model that can be adapted to the unlabeled target domain. To achieve high quality hand localization, we apply an uncertainty-aware domain adaptation network, \ie, MEAA, to train a hand detector, which only uses very limited hand bounding box annotations in the source domain but can generalize well to the unlabelled target domain. The experimental results on the EPIC-KITCHENS-100 dataset demonstrate the effectiveness of our approach for UDA on fine-grained action recognition.

\section{Our Approach}
In this section, we present the technical details of our proposed approach. As illustrated in Fig.~\ref{fig:architect}, the overall architecture has two stages: hand-centric feature generation and video domain adaptation. 

\subsection{Hand-centric Feature Generation}
\label{sec:hcfg}
The hand-centric feature generation stage consists of three key components: feature extractors, domain-adaptive hand detector, and hand-centric feature generator. Next, we will describe each component in details. 

\noindent\textbf{Feature extractors.} To learn discriminative feature representations, we investigate two pre-trained action recognition models for feature extraction, \ie TBN~\cite{kazakos2019TBN} and TSM~\cite{lin2019tsm}. The extracted features are used to generate hand-centric RGB and flow features which serve as the input to the video domain adaptation model. We empirically find that features extracted with TSM model can lead to better domain adaptation results, which is consistent with the action recognition results on EPIC-KITCHENS-100 dataset reported in~\cite{damen2020rescaling}. This indicates that compared with TBN model, TSM model tends to learn more discriminative features for action recognition. Therefore, we employ the features extracted with TSM model in our final submission. 

\begin{figure*}[t]
  \centering
  \includegraphics[width=1.0\linewidth]{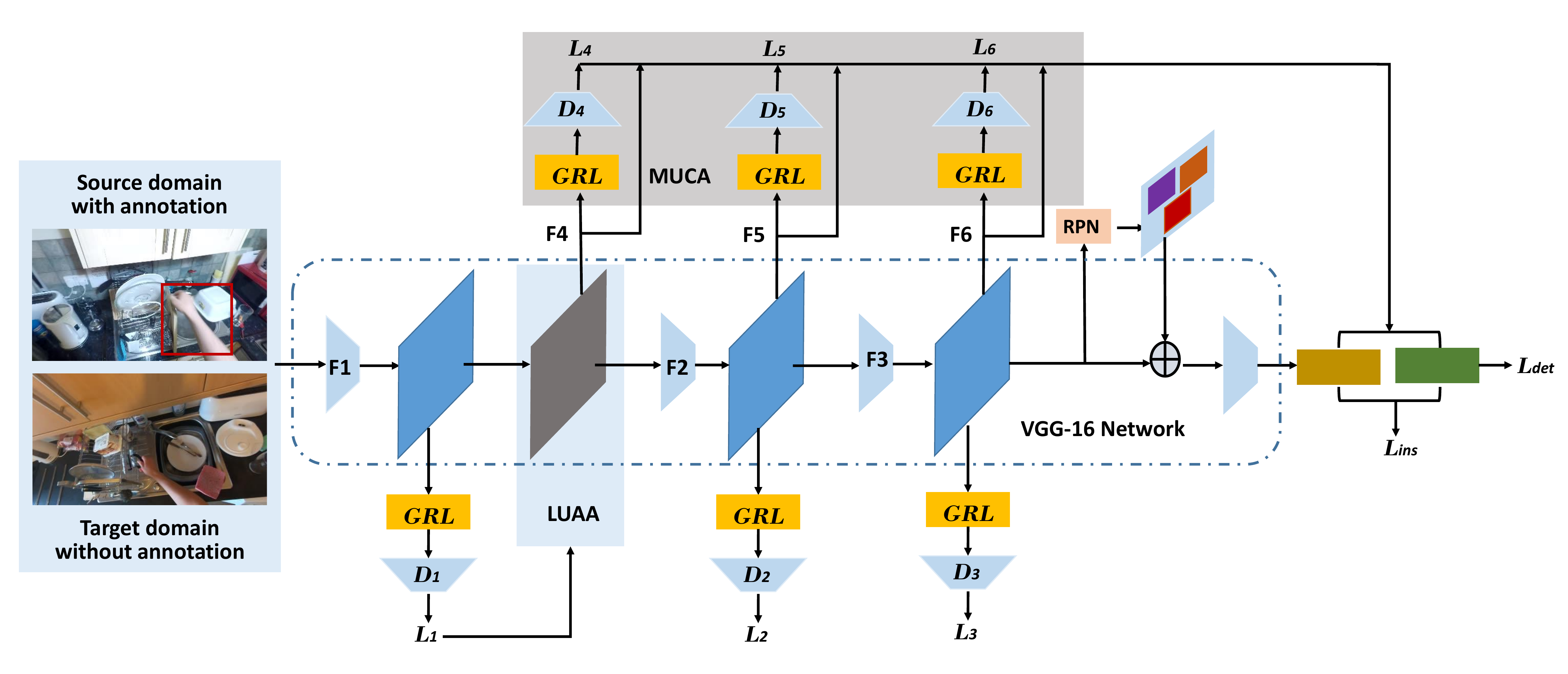}
  \caption{Overall architecture of the domain-adaptive hand detector (MEAA). It consists of two modules, namely LUAA and MUCA. $D_{i}$, $F_{i}$ and $L_{i}$, where $i=1,2,...,6$, denote multi-level domain classifiers, feature extractors and losses, respectively. GRL denotes Gradient Reverse Layer. This is best viewed in color.} 
\label{fig:detector_achitect}
\end{figure*}

\noindent\textbf{Domain-adaptive hand detector.} To achieve high quality hand localization, it is necessary to train a hand detector in labelled source domain and adapt it to unlabelled target domain. This makes the hand detection in our task falls in the area of domain-adaptive object detection which aims to transfer knowledge from labelled source domain to unlabelled target domain. To this end, we adapt the existing method MEAA~\cite{nguyen2020domain} to train a domain-adaptive hand detector, and the detailed structure is presented in Fig.~\ref{fig:detector_achitect}. It investigates (1) uncertainty measurement of input pairs (images in source and target domain) by utilizing domain classifiers at multiple levels of the backbone network as well as (2) uncertainty at image and instance levels to guide the model to pay more attention to hard-to-align instances and images. Specifically, it designs the Local Uncertainty Attentional Alignment (LUAA) module to align high-level features and low-level features by perceiving structure invariant regions of objects and a Multi-level Uncertainty-Aware Context Alignment (MUCA) module to enrich the model with uncertainty-weighted context vectors. 

\noindent\textbf{Hand-centric feature generator.} After obtaining the extracted feature maps and the predicted hand bounding boxes, we generate the hand-centric features to train the domain-adaptive action recognition model. As the RGB and flow features are generated in the same manner, we only describe the steps to generate RGB features. As illustrated in Fig.~\ref{fig:architect}, for each frame, we first extract the RGB features with TSM model pretrained in the source domain and obtain the region of interest (ROI) by applying union operation on all the hand bounding boxes in the frame. Then, the context features are generated by applying the average pooling operation on the extracted RGB feature, while the hand-related features are generated by applying the RoIAlign operation followed by the average pooling operation. By combining the context features and hand-related features with the element-wise addition, we obtain the final hand-centric features. Finally, the hand-centric RGB and flow features are concatenated to serve as the input to the domain-adaptive action recognition model. 

\subsection{Video Domain Adaptation}
\label{sec:vda}
Following the baseline provided by the organizers, we adapt an existing method for video domain adaptation, \i.e., TA3N~\cite{chen2019temporal}, to train the domain-adapted action recognition model. As illustrated in Fig.~\ref{fig:architect}, TA3N designs the temporal relation module to model the $n$-frame temporal relation by taking $n$ temporal-ordered sampled frames as input and output $n$-frame relation features. These relation features are then aggregated to generate the video-level features. Similar to other approaches for video domain adaptation, TA3N applies the adversarial discriminator $\hat{G}_{sd}$ to align the spatial (frame-level) features and the adversarial discriminator $\hat{G}_{td}$ to align the video-level features from different domains. Differently, it designs a set of adversarial discriminators $\hat{G}_{rd}^n$ to align the $n$-frame relation features from different domains. In our solution, we modify the code of TA3N by designing two classifiers for the video-level features, with $G_y^v$ for verb classification and $G_y^n$ for noun classification.

\section{Experiments}

\subsection{Datasets.} The EPIC-KITCHENS-100 dataset~\cite{damen2020rescaling} contains a source domain and a target domain. The source domain contains labelled videos collected in 2018 and the target domain contains unlabelled videos collected in 2020. Videos from both domains are further split into train, valuation and test sets.

\begin{table*}[ht]
\caption{The performance of different models on the EPIC-KITCHENS-100 validation set. ``FeatDim" and ``NumSeg" are hyper-parameters in TA3N, which denote the dimension of shared features and number of input frames, respectively. ``Raw features" denote features extracted with backbone models, while ``Hand-centric features" denote features generated by incorporating the hand bounding box information.}
\vspace{2mm}
\centering
\scalebox{0.9}{
\begin{tabular}{c|c|c|c|c|c|c|c|c|c|c}
\hline
\multirow{2}{*}{Method} & \multirow{2}{*}{Backbone} & \multirow{2}{*}{Input Type} & \multirow{2}{*}{FeatDim} & \multirow{2}{*}{NumSeg} & \multicolumn{3}{c|}{Top-1 Accuracy (\%)} & \multicolumn{3}{c}{Top-5 Accuracy (\%)} \\ \cline{6-11} 
 &  &  &  &  & Verb & Noun & Action & Verb & Noun & Action \\ \hline
TA3N & TBN & Raw features & 512 & 5 & 42.97  &  27.17 & 16.63 &74.34  & 49.01  & 41.31 \\ \hline
TA3N & TSM & Raw features & 512 & 5 & 46.31 & 33.17 & 20.02 & 80.58 & 56.44 & 48.94 \\ \hline
TA3N & TSM & Hand-centric features & 512  & 5  & 48.62 & 35.14 & 21.73 & 80.50 & 57.94 & 50.25\\ \hline
TA3N & TSM & Hand-centric features & 1024 & 20 & 52.37 & 37.00 & 24.48 & 81.13 & 59.18 & 51.75  \\ \hline
\end{tabular}
}
\label{tab:action}
\end{table*}

\begin{table*}[ht]
\caption{The performance of different models on the EPIC-KITCHENS-100 test set. ``Ensemble" denotes whether model ensemble is used to generate the testing results. Other definitions are the same as in Table~\ref{tab:action}.}
\vspace{2mm}
\centering
\scalebox{0.85}{
\begin{tabular}{c|c|c|c|c|c|c|c|c|c|c|c}
\hline
\multirow{2}{*}{Method} & \multirow{2}{*}{Backbone} & \multirow{2}{*}{Input Type} & \multirow{2}{*}{FeatDim} & \multirow{2}{*}{NumSeg} & \multirow{2}{*}{Ensemble} & \multicolumn{3}{c|}{Top-1 Accuracy (\%)} & \multicolumn{3}{c}{Top-5 Accuracy (\%)} \\ \cline{7-12} 
 &  &  &  &  & & Verb & Noun & Action & Verb & Noun & Action \\ \hline
TA3N & TSM & Hand-centric features & 1024 & 20 & No & 52.99 & 34.76 & 24.71 & 80.05 & 58.66 & 40.23 \\\hline
TA3N & TSM & Hand-centric features & 1024 & 20 & Yes & 53.16 & 34.86 & 25.00 & 80.74 & 59.30 & 40.75 \\\hline
\end{tabular}
}
\label{tab:action_test}
\end{table*}

\subsection{Implementation Details}
Following the baseline method in~\cite{damen2020rescaling}, we train our model using a two-stage optimization scheme. Specifically, we first train the TSM~\cite{lin2019tsm} model and hand detection model on source domain data to generate the hand-centric features. Subsequently, we train the modified TA3N~\cite{chen2019temporal} for domain-adaptive action recognition. 

\noindent\textbf{Feature extractors.}
As the organizers provide the RGB and flow features extracted with TBN~\cite{kazakos2019TBN} model pretrained in the source domain, we only need to train TSM~\cite{lin2019tsm} in the source domain for feature extraction. The network parameters are learned with SGD optimizer with momentum $0.9$ and weight decay $5 \times 10^{-4}$. We train the models for 60 epochs, where the learning rate is initialized at $0.01$ and multiplied by $0.1$ for every 20 epochs. The batch size is set at $16$ and the input size is set at $256 \times 256$. During training, we first resize the shorter edge of each frame to 256 while keeping the aspect ratio, and then randomly crop the frame to $256 \times 256$ to feed it into the backbone model. During testing, we take the same resizing strategy but use center crop to generate the input of size $256 \times 256$. After applying the Average Pooling, the dimension of generated feature is $2048$.

\noindent\textbf{Domain-adapted hand detector.}
As shown in Fig.~\ref{fig:detector_achitect}, the training inputs are image pairs (source image with annotation, target image without annotation). 
As no hand bounding box annotation is provided in the source domain, we randomly select a very limited number of frames, \ie, $3100$ images, and annotate the hand bounding boxes manually. 
Meanwhile, we select double size images with hand from target videos. In this case, one source image will appear in two image pairs, and totally we have $6200$ image pairs for training the hand detector. During training, the parameters are learned with Adam optimizer. We train the model for $10$ epochs, and the leaning rate is set as $0.001$ with the decay step as 4.

\noindent\textbf{Video domain adaptation.} 
We follow the guidelines given by the organizers to train the domain-adaptive action recognition model. First, we train the model using the source validation and target validation splits to select the best hyper-parameters. Then, we retrain the model using the source train and target train splits, where the retrained model is evaluated on the target test split to generate the action predictions for this challenge. We adapt TA3N to train our model. During training, the parameters in the feature extractors and the hand detector are freezed. The parameters in domain-adaptive action recognition model are learned using SGD optimizer, where the initial learning rate are set as $3 \times 10^{-3}$. We train the model for $30$ epochs and the learning rate is multiplied by $0.1$ for every 10 epochs. In our submission, the number of input frames and the shared feature dimension are empirically set as $20$ and $1024$, respectively. 


\subsection{Results}
\label{Sec:results}

\begin{figure*}[t]
  \centering
  \includegraphics[width=1.0\linewidth]{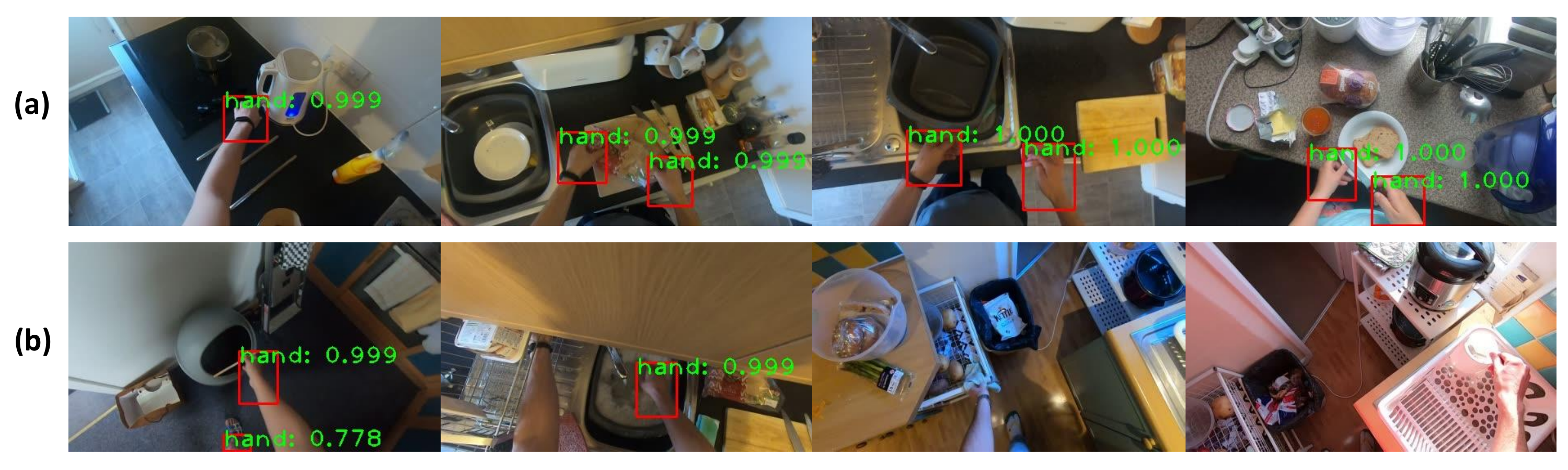}
  \caption{Visualization of hand detection results on unlabelled target domain. The red boxes denote predicted hand bounding boxes generated by the domain-adaptive hand detector. The green characters illustrate the predicted object class of each bounding box and its confidence score. } 
\label{fig:detector_vis}
\end{figure*}

\noindent\textbf{UDA for action recognition.} We employ two backbones (TBN and TSM) trained on the labelled source domain as feature extractors, and train TA3N with different features as inputs. The performance of different models on the validation set are summarized in Table~\ref{tab:action}. All the models use both RGB and optical flow as input. As shown in the table, by replacing the TBN with TSM as feature extractors, the top-1 action accuracy can be improved by 3.30\%. By leveraging the hand bounding box information to generate hand-centric RGB and flow features can further improve the top-1 action accuracy by 1.71\%, where the top-1 noun and verb branches achieve similar performance gains. This demonstrates the effectiveness of the hand-centric features in recognizing active objects as well as understanding the hand motions. More importantly, it also helps to capture the domain-invariant features for accurate action recognition in the target domain. By adapting the hyper-parameters, \ie, shared feature dimension and number of input frames, in TA3N, our final model can achieve 24.48\% in terms of top-1 action accuracy on the validation set. For better robustness, we adopt the same model ensemble strategy as in~\cite{Sun2020TeamVT} to generate our final submission to the challenge based on the best model in Table~\ref{tab:action}. The results on the test set are presented in Table~\ref{tab:action_test}. The best model in Table~\ref{tab:action_test} ranks first in terms of the top-1 action accuracy in the EPIC-KITCHENS-100 UDA Challenge for Action Recognition.

\noindent\textbf{Visualization results of domain-adaptive hand detector.} We present the predicted hand bounding boxes and their confidence scores on selected samples from the unlabeled target domain in Fig.~\ref{fig:detector_vis}. The results demonstrate that the domain-adaptive hand detector trained with very limited labelled samples in the source domain can generalize well to most of the cases in the unlabelled target domain. 
Fig.~\ref{fig:detector_vis} (a) shows that hands can be correctly detected with high confidence scores under normal view angle and lighting condition. Fig.~\ref{fig:detector_vis} (b) shows selected hard samples where there are some false or missed hand detection results. Specifically, false detection may happen when some objects are visually similar as human hands (the first image), while miss detection may happen under heavy occlusions (left hand in the second image), extreme view angle (the third image), or extreme lighting condition (the last image).

\section{Conclusion}
In this report, we describe the technical details of our approach to the EPIC-KITCHENS-100 UDA Challenge for Action Recognition. Specifically, we propose to learn hand-centric features by leveraging the hand bounding box information for UDA on fine-grained action recognition. To obtain high-quality hand localization, we apply MEAA to train a domain-adaptive hand detector with very limited hand bounding boxes annotations in the source domain. The experimental results on the EPIC-KITCHENS-100 dataset demonstrate the effectiveness of our proposed method. With further performance increase from the model ensemble, our final submission ranks first on the leaderboard in terms of top-1 action recognition accuracy.

\section*{Acknowledgments}
We would like to thank Dr. Joo Hwee Lim for his support and guidance. This research is supported by the Agency for Science, Technology and Research (A*STAR) under its AME Programmatic Funding Scheme (Project \#A18A2b0046).

{\small
\bibliographystyle{ieee_fullname}
\bibliography{egbib}
}

\end{document}